%% file: main.tex
\documentclass[letterpaper]{article}

\usepackage[preprint]{aaai2027}

\usepackage[hyphens]{url}
\usepackage{graphicx}
\usepackage{natbib}
\usepackage{caption}
\usepackage{amsmath,amssymb,amsfonts,bm}
\usepackage{booktabs}
\usepackage{multirow}
\newcommand{\Best}[1]{\textbf{#1}}
\newcommand{\Second}[1]{\underline{#1}}

\newcommand{\Rmnum}[1]{\uppercase\expandafter{\romannumeral #1}}

\title{Unleashing the Power of Text: Text-Guided Flow Matching for Image Fusion under Complex Degradations}
\author{
    Axi Niu\textsuperscript{\rm 1},
    Jieheng Li\textsuperscript{\rm 1}\corresponding,
    Kang Zhang\textsuperscript{\rm 2},
    Qingsen Yan\textsuperscript{\rm 1},
    Jinqiu Sun\textsuperscript{\rm 3},
    Yanning Zhang\textsuperscript{\rm 1}
}

\affiliations{
    \textsuperscript{\rm 1} School of Computer Science, Northwestern Polytechnical University, Xi'an, Shaanxi, China\\
    \textsuperscript{\rm 2} School of Electrical Engineering, KAIST, Daejeon, Republic of Korea\\
    \textsuperscript{\rm 3} School of Aeronautics and Astronautics, Northwestern Polytechnical University, Xi'an, Shaanxi, China\\
    hastingslhx@gmail.com
}


\begin{document}

\maketitle

\begin{abstract}
Infrared-visible image fusion under realistic degradation scenarios is a challenging task, as degradations not only cause a loss of reliable modality-specific information in observed images but also hinder the fusion process.
Recent studies indicate that text can provide prior information about degradation characteristics, complementing the limited evidence available from corrupted input images and facilitating fusion.
However, existing methods typically inject fixed global text representations into visual features, making it difficult for textual guidance to adapt to spatially varying degradations, local structures, and thermal saliency.
To this end, we propose TGFusion, a text-guided latent-space flow matching framework that unifies degradation suppression and cross-modal fusion. TGFusion encodes task, degradation, and generation cues into structured prompts. To fully exploit these priors, we design a Prompt-conditioned Multi-stream Joint Flow Transformer that represents text as an independent semantic stream alongside fusion, visible, and infrared streams. Joint attention enables token-level bidirectional interaction and layer-wise updating among semantic and visual representations, allowing degradation semantics to dynamically guide reliable information selection and fusion latent generation. Extensive experiments on public benchmarks and complex degradation scenarios demonstrate that TGFusion achieves superior or competitive performance in perceptual quality, image naturalness, structural-detail preservation, and infrared-saliency retention, while remaining robust across diverse single and compound degradations.

\end{abstract}

\section{Introduction}
Infrared-visible image fusion (IVIF) aims to integrate complementary information from infrared and visible images into a unified representation~\citep{liu2024infrared}. Infrared images emphasize thermally salient targets and remain reliable under illumination variations, whereas visible images provide fine textures and structural details essential for human and machine perception. By combining these modality-specific strengths, IVIF supports a broad range of applications, including surveillance~\citep{zhang2018vehicle}, autonomous driving~\citep{bao2023heat}, object perception~\citep{jain2023multimodal}, and downstream semantic understanding~\citep{zhang2023cmx}.

Deep fusion networks based on CNNs~\citep{liang2022fusion,liu2022target}, autoencoders~\citep{li2018densefuse,li2023lrrnet}, generative models~\citep{ma2019fusiongan}, and Transformers~\citep{ma2022swinfusion,zhang2022transformer} have substantially improved feature extraction, cross-modal interaction, and reconstruction. However, most are developed under relatively ideal imaging conditions and implicitly assume that source images provide reliable modality cues~\citep{zhao2023cddfuse,zhao2024equivariant}. 
In real-world scenarios, infrared and visible images always undergo diverse modality-specific and composite degradations of varying severity, which existing fusion methods often fail to handle, thereby degrading fusion quality and limiting robustness~\citep{yi2024text-if,tang2026controlfusion}.
A straightforward strategy is to restore the degraded input from each modality before fusion. However, this cascaded pipeline incurs additional computational overhead and decouples restoration from fusion, hindering their end-to-end joint optimization~\citep{cao2025mmaif}.

\begin{figure}[!t]
    \centering
    \includegraphics[width=\linewidth]{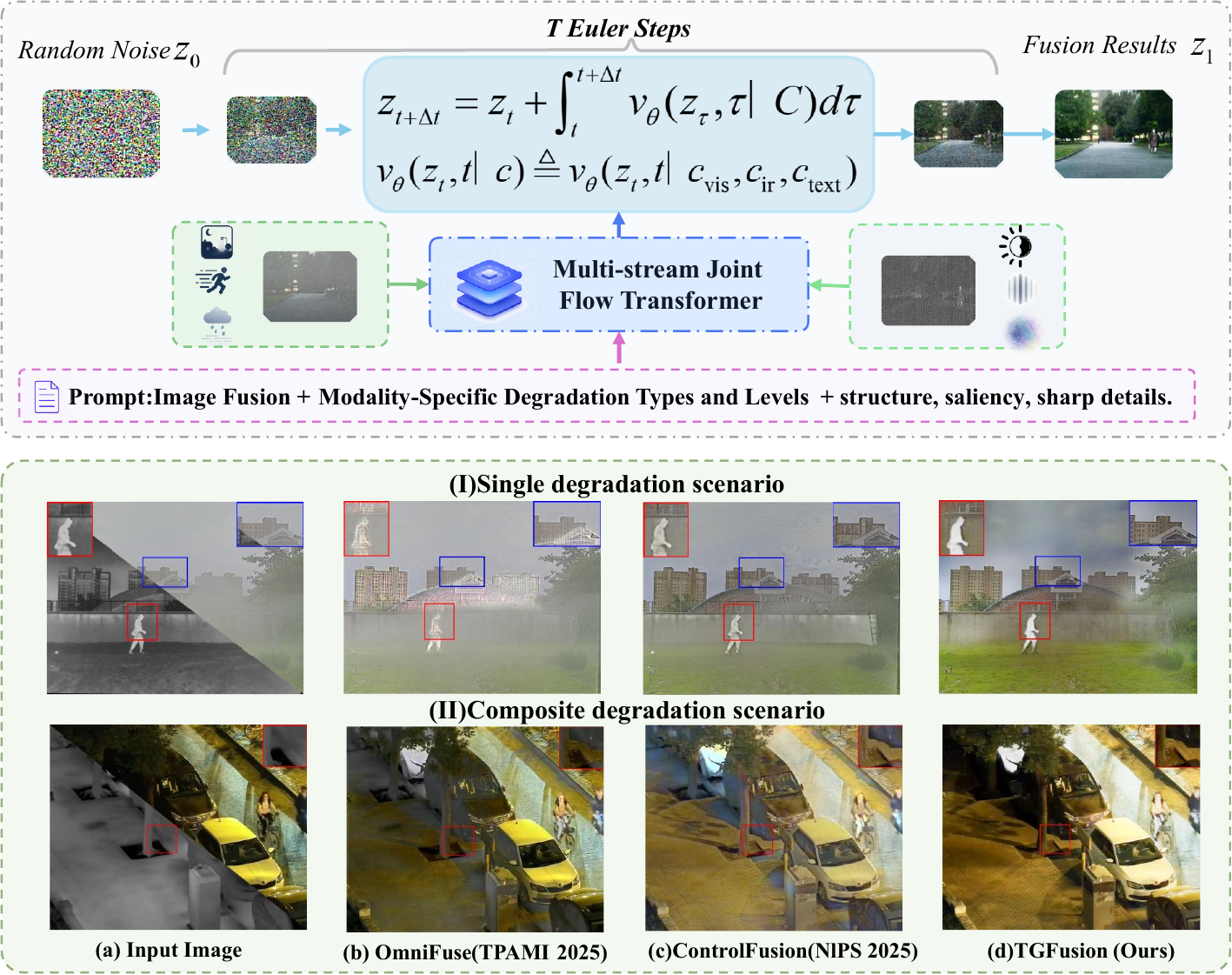}
    \caption{Overview of TGFusion (top) and qualitative comparisons under degradation scenarios (bottom). }
    \label{fig:introduction}
\end{figure}

Recent degradation-aware and text-guided methods encode degradation states or fusion preferences in text and incorporate these semantics into fusion either by modulating visual features with fixed textual representations~\citep{tang2026controlfusion} or by converting them into object-level spatial priors for semantic-aware fusion~\citep{zhang2025omnifuse}. Such textual cues provide information beyond corrupted visual observations and help identify reliable modality-specific evidence.
Despite this benefit, language guidance is generally defined in advance and passed to the visual pathway in a largely one-way manner. Because it is not revised according to the current fusion state, it may not adequately reflect spatially nonuniform corruption or local content variations. Text therefore serves mainly as auxiliary conditioning instead of evolving together with the fused representation during generation. As illustrated in Figure~\ref{fig:introduction}, ControlFusion~\citep{tang2026controlfusion}, which relies on fixed textual conditioning, and OmniFuse~\citep{zhang2025omnifuse}, which translates textual semantics into object-level spatial priors, still leave residual degradation artifacts, obscure local structures, and attenuate infrared targets under degradations. These distortions degrade task-relevant visual cues and can propagate to downstream perception, resulting in erroneous predictions.

Motivated by these limitations, TGFusion formulates IVIF with degraded inputs as conditional transport in a compact latent space, using structured textual descriptions as semantic priors that guide the generative trajectory from noise to the target fusion representation in a compact VAE latent space. Throughout this trajectory, linguistic guidance is continually adapted to the evolving latent state and modality-specific evidence. This progressive semantic--visual interaction links textual priors to scene structures and thermally salient regions, enabling reliable cue selection and fusion representation generation. As shown in Figure~\ref{fig:introduction}, TGFusion produces robust fusion results across diverse degraded conditions. In summary, our main contributions are as follows:

\begin{itemize}
    \item We present TGFusion, which formulates degradation-robust IVIF as text-guided latent-space flow matching and integrates corruption handling with fusion in a single generative process.
    \item We develop a Prompt-conditioned Multi-stream Joint Flow Transformer that organizes text as an independent semantic stream alongside the fusion, visible, and infrared streams. Joint attention supports token-level interaction and layer-wise updating across the four streams, overcoming the limitations of fixed global textual conditioning.
    \item TGFusion consistently outperforms representative two-stage restoration--fusion pipelines and recent all-in-one degradation-aware methods under various degradation conditions. Superior downstream semantic segmentation performance further validates its effectiveness in supporting high-level perception.

\end{itemize}

\section{Related Work}
\label{sec:related}

\begin{figure*}[!t]
    \centering
    \includegraphics[width=0.95\textwidth]{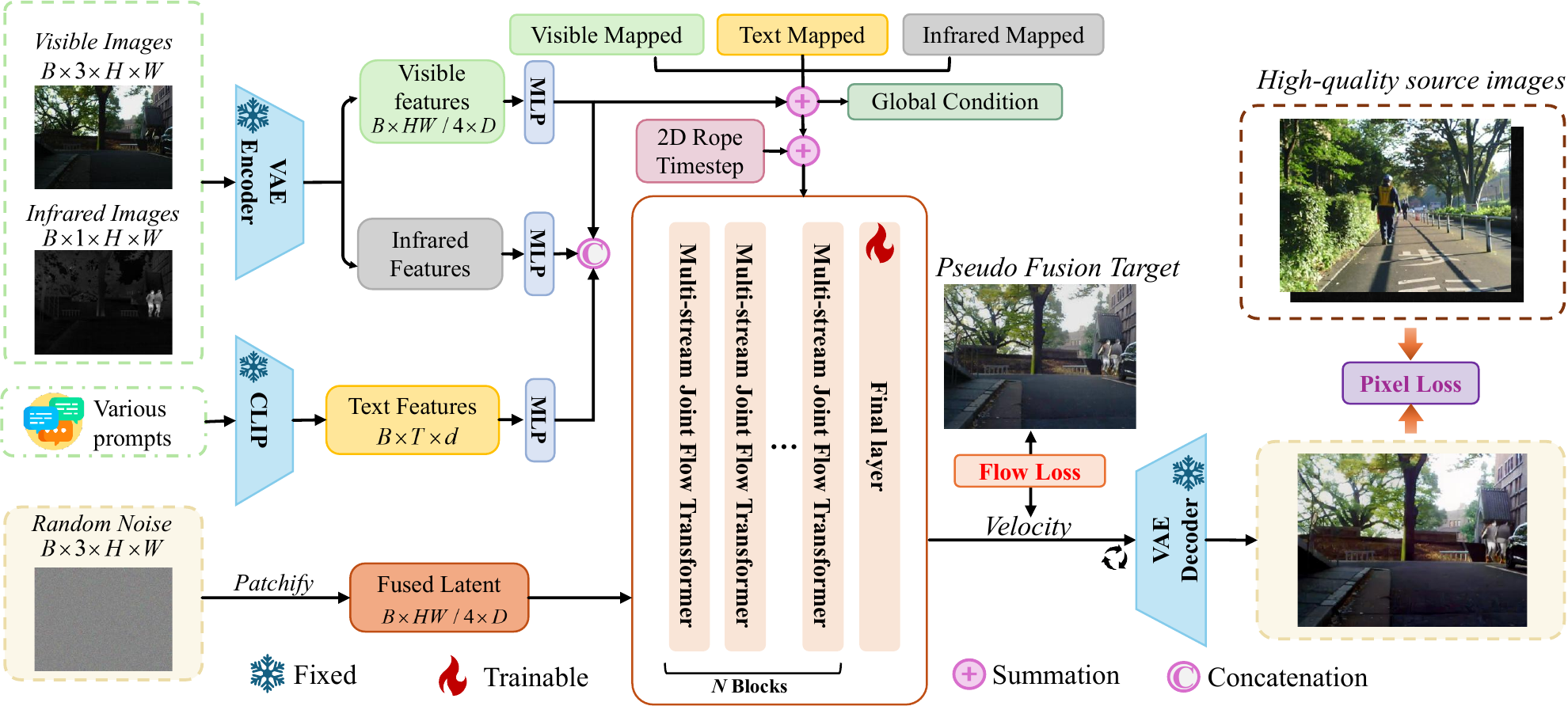}
    \caption{Overview of the proposed TGFusion framework}
    \label{fig:Framework}
\end{figure*}

\begin{figure*}[!t]
    \centering
    \includegraphics[width=0.95\textwidth]{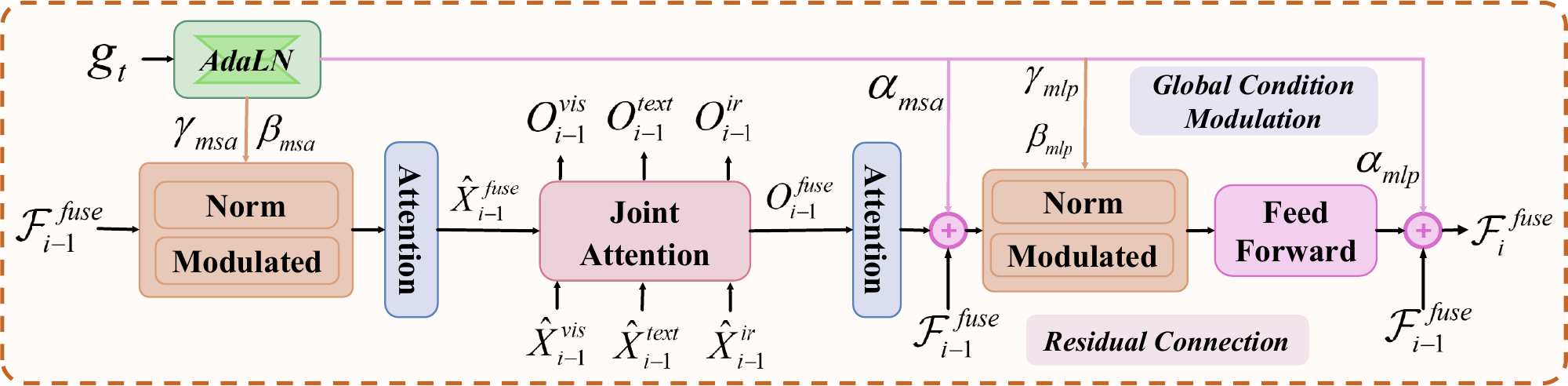}
    \caption{Detailed architecture of the proposed Prompt-conditioned Multi-stream Joint Flow Transformer.}
    \label{fig:Framework2}
\end{figure*}

\subsection{Text-Guided Image Fusion}
Text provides high-level priors that complement visual observations and support degradation-aware or preference-controlled fusion. With vision--language encoders such as CLIP~\citep{radford2021learning}, existing approaches mainly follow two directions. The first describes imaging conditions to improve robustness. Text-IF~\citep{yi2024text-if} uses task and degradation descriptions to regulate restoration and fusion, while ControlFusion~\citep{tang2026controlfusion} introduces language--vision degradation prompts that characterize degradation categories and severities under compound corruptions. The second uses text to specify semantic preferences or regions of interest. Text-DiFuse~\citep{zhang2025text} enhances text-relevant foreground regions through feature remodulation; TextFusion~\citep{cheng2025textfusion} establishes text--region associations for preference-aware fusion; and OmniFuse~\citep{zhang2025omnifuse} combines linguistic semantics with spatial localization for compound-degradation scenarios.

Collectively, these methods establish text as a flexible interface for describing degradation states and fusion preferences. However, its role is generally confined to auxiliary conditioning rather than participating in progressive multimodal integration.

\subsection{Diffusion and Flow Matching for Image Fusion}

Diffusion models formulate image fusion as conditional generation, providing strong priors for structural plausibility and perceptual quality. DDFM~\citep{zhao2023ddfm} combines diffusion posterior sampling with source-image constraints, while CCF~\citep{cao2024conditional} adaptively selects fusion constraints from a condition bank during sampling. For corrupted inputs, DRMF~\citep{tang2024drmf} composes modality-specific degradation priors, extending diffusion fusion to degradation-robust settings. Recent studies move generative fusion into compact latent spaces. OmniFuse~\citep{zhang2025omnifuse} combines latent diffusion with text-driven semantics for compound degradations. MMAIF~\citep{cao2025mmaif} includes a latent-space flow-matching variant~\citep{lipman2023flowmatching} for text-guided fusion across multiple tasks and degradation types.

Overall, generative fusion has progressed from diffusion-based constrained sampling to latent diffusion and flow matching, enabling more flexible modeling of multimodal fusion distributions. Building on this progression, TGFusion adopts latent flow matching as its generative foundation and integrates textual guidance throughout latent evolution.

\section{Method}

\subsection{Preliminaries}

\paragraph{Conditional flow matching.}
Conditional flow matching learns a time-dependent velocity field that transports a tractable noise distribution to a target data distribution~\citep{lipman2023flowmatching}. Let $Z_0\sim\mathcal{N}(0,\mathbf I)$ denote noise and $(Z_1,C)$ denote a target sample with condition $C$. For $t\sim\mathcal U[0,1]$, the linear probability path and its target velocity are
\begin{equation}
    Z_t=(1-t)Z_0+tZ_1,
    \qquad
    u_t=\frac{\mathrm d Z_t}{\mathrm dt}=Z_1-Z_0.
    \label{eq:linear_probability_path}
\end{equation}
The learned conditional velocity field $v_\theta(Z_t,t,C)$ is optimized by
\begin{equation}
\label{eq:flow_loss}
    \mathcal{L}_{\mathrm{flow}}
    =
    \mathbb{E}_{t,Z_0,Z_1,C}
    \left[
    \left\|
    v_\theta(Z_t,t,C)-(Z_1-Z_0)
    \right\|_2^2
    \right].
\end{equation}
At inference, the learned ODE is integrated from $t=0$ to $t=1$ to transform Gaussian noise into a target sample.

\paragraph{Latent-space generative modeling.}
For computational efficiency, a pretrained VAE encoder $\mathcal E$ maps images into compact latent representations, where flow matching based generation is performed, and its decoder $\mathcal D$ reconstructs the generated latent into image space.

\subsection{Overview}

Given a degraded visible image $I_v^d$, a degraded infrared image $I_r^d$, and a structured prompt $p$, TGFusion $\mathcal G_{\theta}$ aims to generate the fused infrared-visible image $\hat{I}_f$
\begin{equation}
    \hat{I}_f
    =
    \mathcal G_{\theta}
    \left(
    Z_0,I_v^d,I_r^d,p
    \right),
    \qquad
    \text{where}~Z_0\sim\mathcal N(0,\mathbf I).
    \label{eq:fusion_mapping}
\end{equation}
Since image fusion has no unique ground-truth target, we use a frozen Text-IF model~\cite{yi2024text-if} to generate a pseudo fusion target $I_f$ from the corresponding clean source pair. As illustrated in Figure~\ref{fig:Framework}, a frozen VAE encodes $I_v^d$, $I_r^d$, and $I_f$ to get the latent-space samples $Z_v$, $Z_r$, and $Z_1$. Meanwhile, a frozen CLIP text encoder converts the prompt $p$ into the semantic condition $c$. The Prompt-conditioned Multi-stream Joint Flow Transformer then predicts the conditional velocity field from $Z_t$, conditioned on $Z_v$, $Z_r$, and $c$.

\subsection{Degradation Semantic Prior Modeling}
\label{sec:semantic_prior}


We construct each prompt from task, modality-specific degradation, and generation-goal descriptions. The degradation description specifies the types and severities of the source images.
ChatGPT~\citep{chatgpt} is used to generate diverse linguistic variants. The prompt is formulated as
\begin{equation}
    p=q_{\mathrm{task}}\oplus q_{\mathrm{deg}}\oplus q_{\mathrm{goal}},
\end{equation}
where $\oplus$ denotes textual concatenation. During training, prompt variants consistent with the annotated degradation states are randomly sampled, while fixed variants are used during inference for reproducibility.

A frozen CLIP~\citep{radford2021learning} text encoder $\tau(\cdot)$ maps $p$ into token-level representations:
\begin{equation}
    c=\tau(p)=[c_1,c_2,\ldots,c_L],
\end{equation}
where $L$ denotes the token-sequence length. The complete token sequence is preserved as the textual input to the subsequent flow Transformer.

\subsection{Prompt-conditioned Multi-stream Joint Flow Transformer}
\label{sec:multi_stream_transformer}

The Prompt-conditioned Multi-stream Joint Flow Transformer parameterizes the flow dynamics.
As shown in Figure~\ref{fig:Framework2}, the proposed transformer preserves stream-specific representations, aggregates the timestep embedding and condition summaries into a global condition $g_t$, and progressively exchanges information across streams through joint attention, stream-wise AdaLN, and gated residual updates. The refined fusion stream is then projected to $\hat{u}_t$.


\paragraph{Multi-stream input embedding and global-condition construction.}
Let $\mathcal M=\{\mathrm{fuse},\mathrm{vis},\mathrm{ir},\mathrm{text}\}$ denote the four streams. The visual latents $Z_t$, $Z_v$, and $Z_r$ are flattened into spatial token sequences, and $c$ is the CLIP token sequence for global semantic prior. Stream-specific projections $\Pi^{\mathrm{fuse}}$, $\Pi^{\mathrm{vis}}$, $\Pi^{\mathrm{ir}}$, and $\Pi^{\mathrm{text}}$ map them to the initial representations $F_0^{\mathrm{fuse}}$, $F_0^{\mathrm{vis}}$, $F_0^{\mathrm{ir}}$, and $F_0^{\mathrm{text}}$, respectively, within a shared hidden space. The fusion stream represents the generative state, while the remaining streams provide visual and semantic conditions.
A global condition vector is constructed as
\begin{equation}
    g_t
    =
    \psi(t)
    +
    \operatorname{MLP}
    \left(
    \sum_{m\in\{\mathrm{vis},\mathrm{ir},\mathrm{text}\}}
    P^m\rho(F_0^m)
    \right),
    \label{eq:global_condition}
\end{equation}
where $\psi(t)$ is the time embedding, $\rho(\cdot)$ is token-wise average pooling, and $P^m$ is a stream-specific projection. Pooling is used only for AdaLN conditioning; the complete token sequences remain available to joint attention.

\paragraph{Prompt-conditioned joint block.}
For each block $l\in\{0,\ldots,L_b-1\}$, the global condition $g_t$ is mapped to stream-specific AdaLN~\citep{peebles2023scalable} parameters:
\begin{equation}
(b_{l,a}^{m},s_{l,a}^{m},\gamma_{l,a}^{m},
 b_{l,f}^{m},s_{l,f}^{m},\gamma_{l,f}^{m})
=\Gamma_l^m(g_t),\quad m\in\mathcal M .
\label{eq:adaln_params}
\end{equation}
After AdaLN modulation, each stream produces queries, keys, and values. Two-dimensional rotary position encoding~\citep{su2024roformer} is applied to the fusion, visible, and infrared streams, while the text stream retains the positional information encoded by CLIP. The resulting QKV representations are concatenated along the token dimension as $Q_l$, $K_l$, and $V_l$, and joint attention is computed by
\begin{equation}
O_l=
\operatorname{Softmax}
\left(
\frac{Q_lK_l^\top}{\sqrt{d_h}}
\right)V_l ,
\label{eq:joint_attention}
\end{equation}
where $d_h$ is the head dimension. The output $O_l$ is subsequently split according to the original token ranges, and each stream is updated through its own output projection, gated residual connection, and FFN. This joint operation enables layer-wise bidirectional interaction among semantic and visual representations while preserving stream-specific transformation paths.

\paragraph{Conditional velocity-field prediction.}
After $L_b$ blocks, the fusion stream predicts the conditional velocity:
\begin{equation}
    \hat{u}_t
    =
    W_o
    \left(
    (1+s_o)\odot \operatorname{LN}(F_{L_b}^{\mathrm{fuse}})
    +
    b_o
    \right),
    \label{eq:velocity_head}
\end{equation}
where $(b_o,s_o)$ are generated from $g_t$, and $W_o$ restores the latent spatial layout. The remaining streams serve only as jointly updated conditional contexts and have no separate prediction heads.

\begin{table*}[t]
    \centering
    \renewcommand{\arraystretch}{1.05}
    {
    \fontsize{9}{10.5}\selectfont
    \setlength{\tabcolsep}{1pt}
    \begin{tabular}{@{}l*{18}{c}@{}}
        \toprule
        & \multicolumn{6}{c}{\textbf{MSRS}}
        & \multicolumn{6}{c}{\textbf{LLVIP}}
        & \multicolumn{6}{c}{\textbf{M3FD}} \\
        \cmidrule(lr){2-7} \cmidrule(lr){8-13} \cmidrule(lr){14-19}
        \multirow{-2}{*}{\textbf{Methods}}
        & \textbf{EN} & \textbf{AG} & \textbf{CC} & \textbf{CLIP-IQA} & \textbf{TReS} & \textbf{NIQE}
        & \textbf{EN} & \textbf{AG} & \textbf{CC} & \textbf{CLIP-IQA} & \textbf{TReS} & \textbf{NIQE}
        & \textbf{EN} & \textbf{AG} & \textbf{CC} & \textbf{CLIP-IQA} & \textbf{TReS} & \textbf{NIQE} \\
        \midrule
        \textbf{CDDFuse}
        & 6.70 & 3.73 & \Best{0.60} & 0.14 & 31.24 & \Second{3.92}
        & 7.35 & 4.30 & \Best{0.69} & 0.37 & 57.05 & 4.11
        & 6.90 & 4.86 & \Best{0.54} & 0.29 & 61.16 & 4.22 \\
        \textbf{EMMA}
        & 6.72 & 3.78 & \Second{0.60} & 0.18 & 27.01 & 4.57
        & 7.35 & 4.67 & \Second{0.69} & 0.40 & 51.93 & 4.23
        & 6.92 & \Second{5.33} & \Second{0.50} & \Best{0.40} & 55.64 & 5.45 \\
        \textbf{Text-IF}
        & 6.72 & 3.82 & 0.59 & 0.16 & 32.55 & 3.97
        & 7.33 & 5.06 & 0.67 & 0.38 & 58.86 & 3.67
        & 6.84 & 5.04 & 0.47 & 0.26 & 62.52 & 4.41 \\
        \textbf{Text-DiFuse}
        & 7.23 & \Second{4.15} & 0.59 & 0.13 & 33.26 & 5.17
        & \Best{7.67} & \Second{5.36} & 0.66 & 0.33 & 61.19 & 4.38
        & \Second{7.02} & 4.68 & 0.48 & 0.31 & 63.28 & 4.68 \\
        \textbf{DRMF}
        & \Second{7.27} & 4.08 & 0.52 & 0.16 & \Second{39.13} & 4.60
        & 7.41 & \Best{5.99} & 0.55 & \Second{0.43} & 66.83 & \Second{3.60}
        & 6.74 & 4.51 & 0.43 & 0.33 & 68.76 & 4.77 \\
        \textbf{OmniFuse}
        & 7.06 & 3.36 & 0.54 & \Second{0.20} & 32.52 & 6.28
        & 7.26 & 3.72 & 0.67 & 0.42 & 49.02 & 5.50
        & 6.88 & 4.78 & 0.47 & 0.30 & 49.61 & 5.95 \\
        \textbf{ControlFusion}
        & 7.11 & 3.91 & 0.55 & 0.14 & 36.43 & 3.95
        & 7.36 & 5.02 & 0.68 & 0.36 & \Second{68.03} & 3.95
        & 6.83 & 4.60 & 0.50 & 0.30 & \Best{72.13} & \Best{3.85} \\
        \textbf{TGFusion}
        & \Best{7.49} & \Best{4.16} & 0.47 & \Best{0.28} & \Best{65.70} & \Best{3.67}
        & \Second{7.51} & 4.55 & 0.61 & \Best{0.48} & \Best{69.92} & \Best{3.27}
        & \Best{7.23} & \Best{5.63} & 0.40 & \Second{0.37} & \Second{69.82} & \Second{4.15} \\
        \bottomrule
    \end{tabular}
    }
    \caption{Quantitative comparison on the MSRS, LLVIP, and M3FD datasets. Complete TNO results are reported in Appendix~\ref{app:quantitative_tno}.}
    \label{tab:normal}
\end{table*}

\begin{table*}[!t]
    \centering
    \renewcommand \arraystretch{1.05}
    {
    \fontsize{8}{9.5}\selectfont
    \setlength{\tabcolsep}{1pt}
        \begin{tabular}{@{}l*{4}{c}@{\hspace{2pt}}*{4}{c}@{\hspace{2pt}}*{4}{c}@{\hspace{2pt}}*{4}{c}@{}}
            \toprule
             & \multicolumn{4}{c}{\textbf{VI (Random noise, RN)}} & \multicolumn{4}{c}{\textbf{VI (Over-exposure, OE)}} & \multicolumn{4}{c}{\textbf{VI (Blur)}} & \multicolumn{4}{c}{\textbf{VI (Low-light, LL)}} \\
            \cmidrule(lr){2-5} \cmidrule(lr){6-9} \cmidrule(lr){10-13} \cmidrule(lr){14-17}
            \multirow{-2}{*}{\textbf{Methods}} & \textbf{CLIP-IQA} & \textbf{MUSIQ} & \textbf{TReS} & \textbf{EN} & \textbf{CLIP-IQA} & \textbf{MUSIQ} & \textbf{TReS} & \textbf{EN} & \textbf{CLIP-IQA} & \textbf{MUSIQ} & \textbf{TReS} & \textbf{EN} & \textbf{CLIP-IQA} & \textbf{MUSIQ} & \textbf{TReS} & \textbf{EN}\\
            \textbf{CDDFuse} & 0.27 & 45.27 & 61.34 & 7.38 & 0.15 & 41.24 & 44.29 & \Second{7.52} & 0.14 & 34.71 & 33.03 & 7.34 & 0.11 & 42.33 & 42.16 & 6.80\\
            \textbf{EMMA} & 0.23 & 42.51 & 46.14 & 7.41 & 0.14 & 39.40 & 38.27 & \Best{7.56} & 0.14 & 33.90 & 27.50 & 7.38 & 0.16 & 41.53 & 34.93 & 6.82\\
            \textbf{Text-IF} & 0.26 & 44.76 & 60.54 & 7.40 & 0.13 & 41.38 & 48.39 & 7.15 & 0.16 & 34.67 & 34.16 & 7.36 & 0.14 & 39.97 & 44.42 & 7.17\\
            \textbf{Text-DiFuse} & 0.22 & 42.01 & 50.99 & \Second{7.51} & 0.12 & 37.97 & 39.75 & 7.45 & 0.15 & 35.01 & 35.14 & \Second{7.51} & 0.16 & 42.36 & 41.57 & 7.22\\
            \textbf{DRMF} & \Second{0.27} & 47.02 & \Second{67.68} & 7.49 & 0.18 & 42.91 & \Second{53.39} & 7.24 & 0.23 & \Second{35.06} & \Second{39.84} & 7.46 & 0.22 & 44.40 & \Second{51.00} & \Second{7.36}\\
            \textbf{OmniFuse} & 0.12 & 38.09 & 38.11 & 7.27 & \Second{0.25} & 37.70 & 41.08 & 7.34 & \Second{0.24} & 32.58 & 32.44 & 7.25 & \Second{0.24} & 39.75 & 37.05 & 7.03\\
            \textbf{ControlFusion} & 0.19 & \Second{47.60} & 58.02 & 7.28 & 0.16 & \Second{45.94} & 52.18 & 7.32 & 0.14 & 31.86 & 31.15 & 7.21 & 0.15 & \Second{44.79} & 48.62 & 6.87\\
            \textbf{TGFusion} & \Best{0.30} & \Best{57.40} & \Best{83.73} & \Best{7.52} & \Best{0.26} & \Best{49.90} & \Best{69.60} & 7.52 & \Best{0.29} & \Best{60.74} & \Best{87.77} & \Best{7.60} & \Best{0.24} & \Best{51.42} & \Best{72.78} & \Best{7.46}\\
            \midrule
             & \multicolumn{4}{c}{\textbf{VI (Rain)}} & \multicolumn{4}{c}{\textbf{VI (Rain and Haze, RH)}} & \multicolumn{4}{c}{\textbf{IR (Random noise, RN)}} & \multicolumn{4}{c}{\textbf{IR (Low-contrast, LC)}} \\
            \cmidrule(lr){2-5} \cmidrule(lr){6-9} \cmidrule(lr){10-13} \cmidrule(lr){14-17}
            \multirow{-2}{*}{\textbf{Methods}} & \textbf{CLIP-IQA} & \textbf{MUSIQ} & \textbf{TReS} & \textbf{SD} & \textbf{CLIP-IQA} & \textbf{MUSIQ} & \textbf{TReS} & \textbf{EN} & \textbf{CLIP-IQA} & \textbf{MUSIQ} & \textbf{TReS} & \textbf{EN} & \textbf{CLIP-IQA} & \textbf{MUSIQ} & \textbf{TReS} & \textbf{EN}\\
            \textbf{CDDFuse} & 0.14 & 43.36 & 43.62 & 47.11 & 0.37 & 59.51 & 71.68 & 6.99 & \Second{0.25} & 42.00 & 41.32 & 7.40 & 0.20 & 45.61 & 47.30 & 6.87\\
            \textbf{EMMA} & 0.15 & 41.90 & 36.25 & 50.50 & \Best{0.46} & 58.31 & 58.98 & 7.07 & 0.21 & 39.00 & 35.00 & 7.40 & 0.18 & 44.84 & 41.28 & 7.20\\
            \textbf{Text-IF} & 0.17 & 43.86 & 44.65 & 46.41 & 0.34 & 60.91 & 71.99 & 6.81 & 0.20 & 31.44 & 45.21 & 7.35 & 0.15 & 42.39 & 47.81 & 7.15\\
            \textbf{Text-DiFuse} & 0.13 & 39.96 & 39.82 & \Second{56.63} & 0.36 & 60.36 & 64.49 & \Second{7.25} & 0.16 & \Second{48.64} & \Second{56.69} & \Second{7.51} & 0.13 & 44.08 & 47.07 & 6.94\\
            \textbf{DRMF} & 0.20 & 44.60 & \Second{48.86} & 52.03 & \Second{0.43} & \Best{65.82} & 76.29 & 7.15 & 0.20 & 45.07 & 53.71 & 7.40 & 0.20 & 44.34 & 52.76 & \Second{7.46}\\
            \textbf{OmniFuse} & \Best{0.25} & 38.30 & 38.00 & 41.36 & 0.36 & 48.26 & 44.79 & 6.76 & 0.23 & 39.09 & 38.74 & 7.26 & \Second{0.24} & 40.06 & 39.53 & 7.26\\
            \textbf{ControlFusion} & 0.16 & \Best{44.83} & 47.88 & 41.94 & 0.36 & 63.20 & \Best{83.11} & 6.84 & 0.16 & 44.73 & 49.08 & 7.30 & 0.16 & \Second{46.92} & \Second{59.33} & 7.18\\
            \textbf{TGFusion} & \Second{0.23} & \Second{44.70} & \Best{56.75} & \Best{58.06} & 0.41 & \Second{64.75} & \Second{82.05} & \Best{7.35} & \Best{0.25} & \Best{52.07} & \Best{76.43} & \Best{7.56} & \Best{0.25} & \Best{51.47} & \Best{76.21} & \Best{7.53}\\
            \midrule
             & \multicolumn{4}{c}{\textbf{IR (Stripe noise, SN)}} & \multicolumn{4}{c}{\textbf{VI (LL) and IR (SN)}} & \multicolumn{4}{c}{\textbf{VI (RH) and IR (RN)}} & \multicolumn{4}{c}{\textbf{VI (OE) and IR (LC)}} \\
            \cmidrule(lr){2-5} \cmidrule(lr){6-9} \cmidrule(lr){10-13} \cmidrule(lr){14-17}
            \multirow{-2}{*}{\textbf{Methods}} & \textbf{CLIP-IQA} & \textbf{MUSIQ} & \textbf{TReS} & \textbf{EN} & \textbf{CLIP-IQA} & \textbf{MUSIQ} & \textbf{TReS} & \textbf{EN} & \textbf{CLIP-IQA} & \textbf{MUSIQ} & \textbf{TReS} & \textbf{EN} & \textbf{CLIP-IQA} & \textbf{MUSIQ} & \textbf{TReS} & \textbf{EN}\\
            \textbf{CDDFuse} & 0.20 & 44.36 & 45.78 & 7.34 & 0.28 & 53.55 & 62.97 & 6.93 & \Best{0.52} & 53.64 & 65.89 & 7.05 & 0.35 & 59.75 & 76.27 & \Best{7.51}\\
            \textbf{EMMA} & 0.18 & 43.01 & 39.24 & 7.39 & 0.36 & 52.76 & 54.76 & 6.96 & 0.45 & 53.70 & 65.80 & 7.04 & 0.30 & 56.67 & 63.67 & 7.38\\
            \textbf{Text-IF} & 0.19 & 44.13 & 46.43 & 7.37 & 0.25 & 58.11 & 72.50 & 7.07 & \Second{0.46} & 53.59 & 76.46 & 7.00 & 0.29 & 58.61 & 77.71 & 6.95\\
            \textbf{Text-DiFuse} & 0.15 & 42.15 & 43.69 & \Second{7.50} & 0.26 & 58.69 & 63.23 & \Second{7.44} & 0.45 & 60.85 & 76.41 & \Second{7.29} & 0.30 & 56.37 & 72.60 & 7.21\\
            \textbf{DRMF} & 0.20 & \Second{45.08} & \Second{54.09} & 7.42 & \Second{0.39} & \Second{62.03} & \Second{75.64} & 7.39 & 0.43 & \Best{65.02} & 74.40 & 7.11 & \Second{0.35} & 61.50 & 79.07 & 7.09\\
            \textbf{OmniFuse} & \Second{0.24} & 39.47 & 39.32 & 7.26 & 0.32 & 49.76 & 49.32 & 7.05 & 0.35 & 44.61 & 43.94 & 6.76 & 0.32 & 50.91 & 56.57 & 7.03\\
            \textbf{ControlFusion} & 0.17 & 44.50 & 50.31 & 7.31 & 0.33 & 59.22 & 71.55 & 7.06 & 0.36 & 64.08 & \Best{82.57} & 6.88 & 0.33 & \Second{64.63} & \Best{83.25} & 7.27\\
            \textbf{TGFusion} & \Best{0.25} & \Best{53.47} & \Best{80.60} & \Best{7.60} & \Best{0.41} & \Best{63.96} & \Best{82.51} & \Best{7.48} & 0.41 & \Second{64.71} & \Second{82.04} & \Best{7.32} & \Best{0.41} & \Best{65.05} & \Second{81.91} & \Second{7.39}\\
            \bottomrule
        \end{tabular}
    }
    \caption{Quantitative comparisons on the DDL-12 benchmark under complex degradation scenarios.}
    \label{tab:degradation}
\end{table*}

\subsection{Optimization and Inference}
\label{Sec:loss}

TGFusion combines the latent-space flow matching objective in Eq.~\ref{eq:flow_loss} with pixel-space fusion regularization. Given the current latent state $Z_t$, the predicted velocity $\hat u_t=v_\theta(Z_t,t,Z_v,Z_r,c)$ is extrapolated over the remaining interval $1-t$ to estimate the final fusion latent at $t=1$, which is then decoded into the fused image:
\begin{equation}
    \hat{Z}_1=Z_t+(1-t)\hat{u}_t,
    \qquad
    \hat{I}_f=\mathcal{D}(\hat{Z}_1),
    \label{eq:predicted_fusion_image}
\end{equation}
where $\hat Z_1$ denotes the estimated fusion latent at the endpoint of the flow trajectory, and $\mathcal D(\cdot)$ is the frozen VAE decoder. Following widely used fusion constraints~\citep{tang2022piafusion,yi2024text-if}, intensity and maximum-gradient terms are imposed on the luminance channel:
\begin{equation}
    \mathcal{L}_{\mathrm{int}}
    =
    \frac{1}{HW}
    \left\|
    \mathcal Y(\hat{I}_f)
    -\max\!\left(\mathcal Y(I_v^c),\mathcal Y(I_r^c)\right)
    \right\|_1,
    \label{eq:intensity_loss}
\end{equation}
\begin{equation}
    \mathcal{L}_{\mathrm{grad}}
    =
    \frac{1}{HW}
    \left\|
    |\nabla \mathcal Y(\hat{I}_f)|
    -
    \max\!\left(
    |\nabla\mathcal Y(I_v^c)|,
    |\nabla\mathcal Y(I_r^c)|
    \right)
    \right\|_1,
    \label{eq:gradient_loss}
\end{equation}
where $I_v^c$ and $I_r^c$ are clean source images, $\mathcal Y(\cdot)$ extracts luminance, and $\nabla$ is the Sobel operator. The objective is
\begin{equation}
    \mathcal{L}_{\mathrm{pixel}}
    =
    \mathcal{L}_{\mathrm{int}}
    +
    \lambda_g\mathcal{L}_{\mathrm{grad}},
    \label{eq:pixel_loss}
\end{equation}
\begin{equation}
    \mathcal{L}_{\mathrm{total}}
    =
    \mathcal{L}_{\mathrm{flow}}
    +
    \lambda_{\mathrm{pix}}\mathcal{L}_{\mathrm{pixel}},
    \label{eq:total_loss}
\end{equation}
where $\lambda_g$ and $\lambda_{\mathrm{pix}}$ are fixed weighting coefficients.

At inference, starting from $Z_0\sim\mathcal N(0,\mathbf I)$, the conditional ODE is solved with Euler updates
\begin{equation}
    Z_{i+1}=Z_i+(t_{i+1}-t_i)
    v_\theta(Z_i,t_i,Z_v,Z_r,c),
\end{equation}
and the final result is decoded as $\hat I_f=\mathcal D(Z_N)$.

\begin{figure*}[!t]
    \centering
    \includegraphics[width=0.95\textwidth]{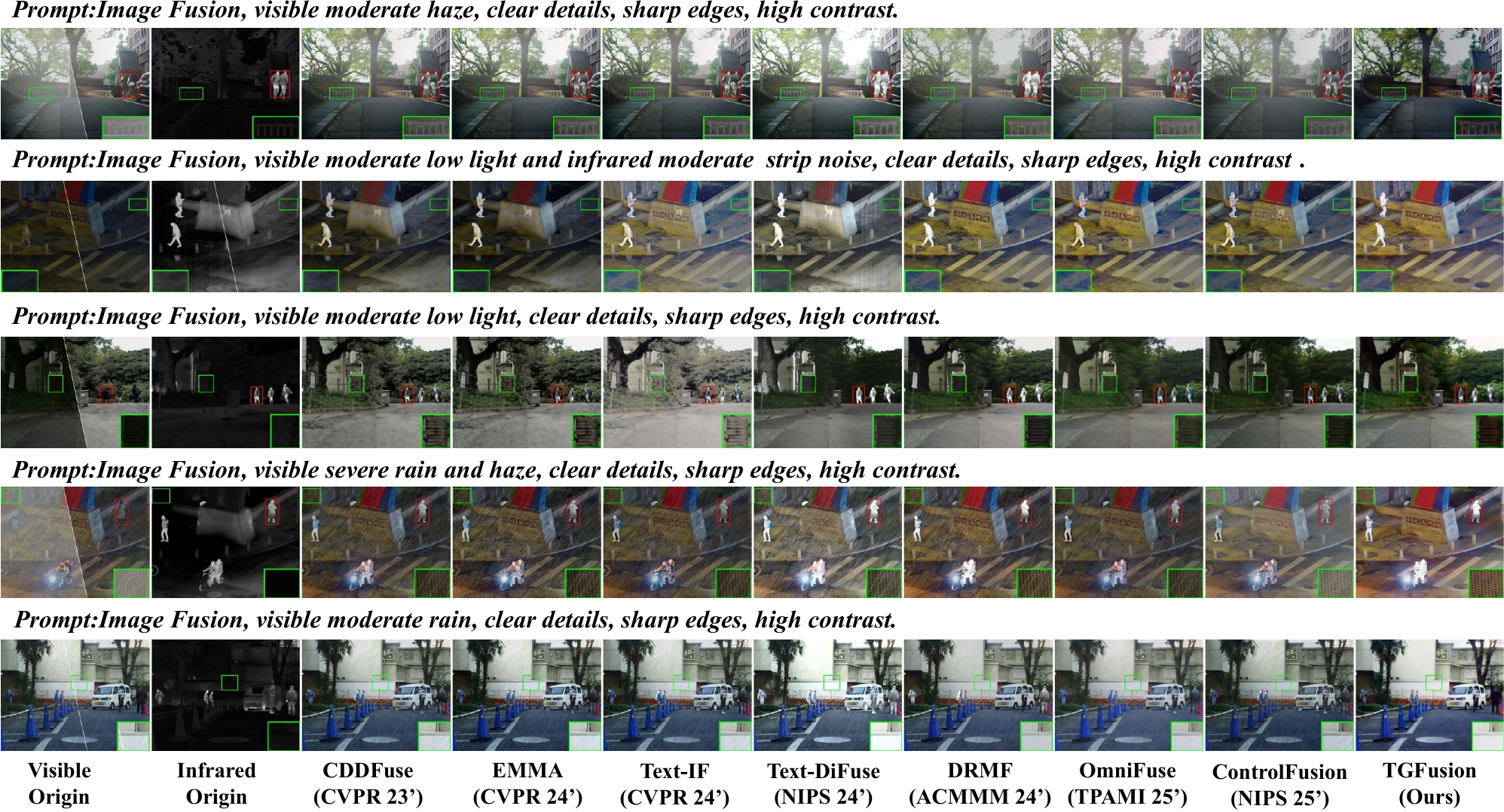}
    \caption{Qualitative comparison of fused results under complex degradation scenarios of the DDL-12 degradation benchmark. }
    \label{fig:degradation}
\end{figure*}

\section{Experiments}

\subsection{Experimental Setup}
\label{sec:experimental_setup}
\noindent\textbf{Training details.}
TGFusion adopts a frozen Stable Diffusion 3 VAE~\citep{esser2024scaling} and a frozen CLIP ViT-L/14 text encoder~\citep{radford2021learning}. Its backbone consists of 12 Prompt-conditioned Multi-stream Joint Flow Transformer blocks, with a hidden dimension of 768 and 12 attention heads. Training is conducted on DDL-12~\citep{tang2026controlfusion} using AdamW~\citep{loshchilov2019decoupled} with a peak learning rate of $1\times10^{-4}$. The loss weights are set to $\lambda_{\mathrm{pix}}=4$ and $\lambda_g=3$. Further implementation and optimization details are provided in Appendix~\ref{app:training_details}.

\noindent\textbf{Evaluation datasets.}
The evaluation covers four public IVIF benchmarks: MSRS~\citep{tang2022piafusion}, LLVIP~\citep{jia2021llvip}, M3FD~\citep{liu2022target}, and TNO~\citep{toet2017tno}. Robustness to input degradations is assessed on the DDL-12 benchmark~\citep{tang2026controlfusion}.

\noindent\textbf{Comparison methods.}
Comparison methods include CDDFuse~\citep{zhao2023cddfuse}, EMMA~\citep{zhao2024equivariant}, Text-IF~\citep{yi2024text-if}, Text-DiFuse~\citep{zhang2025text}, DRMF~\citep{tang2024drmf}, OmniFuse~\citep{zhang2025omnifuse}, and ControlFusion~\citep{tang2026controlfusion}. For methods requiring pre-enhancement, we use SwinIR~\citep{liang2021swinir}, AirNet~\citep{airnet}, and WDNN~\citep{guan2019wavelet} for infrared noise, low contrast, and stripe noise, respectively; LMPEC~\citep{afifi2021learning} for visible overexposure; and AdaIR~\citep{cui2025adair} for the remaining visible degradations. 

\noindent\textbf{Metrics.}
Since source-referenced metrics can be confounded by degradations in the input images~\citep{zhang2024mrfs}, we combine non-reference measures of information and detail (EN, SD, and AG), perceptual-quality metrics (CLIP-IQA, MUSIQ, TReS, and NIQE), and CC as a complementary measure of source fidelity. 

\subsection{Fusion Performance on Public Benchmarks}
\label{sec:public_benchmarks}

Table~\ref{tab:normal} presents quantitative comparisons on public benchmarks. TGFusion achieves the best or second-best results in nearly all non-reference evaluations, demonstrating strong information and detail preservation, perceptual quality, and naturalness. Despite these advantages, TGFusion obtains only moderate CC scores, whereas CDDFuse and EMMA, which do not explicitly model input degradations, achieve higher CC without consistent gains in non-reference metrics. Since public benchmarks contain modality-specific degradations~\citep{zhang2025omnifuse}, source correlation reflects content retention but cannot distinguish informative structures from degraded content. The qualitative comparisons in Appendix~\ref{app:visual_public} further show that higher-CC results also retain visible degradations and artifacts. We therefore primarily use no-reference metrics in subsequent comparisons.

\begin{table*}[!t]
    \centering
    \renewcommand{\arraystretch}{1.05}
    \setlength{\tabcolsep}{2.5pt}
    {\small
    \begin{tabular}{@{}l*{10}{c}@{}}
        \toprule
        \textbf{Metric} & \textbf{Infrared} & \textbf{Visible} & \textbf{CDDFuse} & \textbf{EMMA} & \textbf{Text-IF} & \textbf{Text-DiFuse} & \textbf{DRMF} & \textbf{OmniFuse} & \textbf{ControlFusion} & \textbf{TGFusion} \\
        \midrule
        \textbf{mIoU} & 65.53 & 69.98 & 70.85 & 70.63 & 70.72 & \Second{71.75} & 70.04 & 69.12 & 71.10 & \Best{71.83} \\
        \textbf{mAcc} & 75.14 & 80.30 & 79.17 & 78.82 & 80.25 & 80.84 & 79.82 & 78.49 & \Second{81.08} & \Best{81.47} \\
        \bottomrule
    \end{tabular}
    }
    \caption{Semantic segmentation evaluation on MFNet.}
    \label{tab:semantic-segmentation}
\end{table*}

\begin{figure}[!t]
    \centering
    \includegraphics[width=\linewidth]{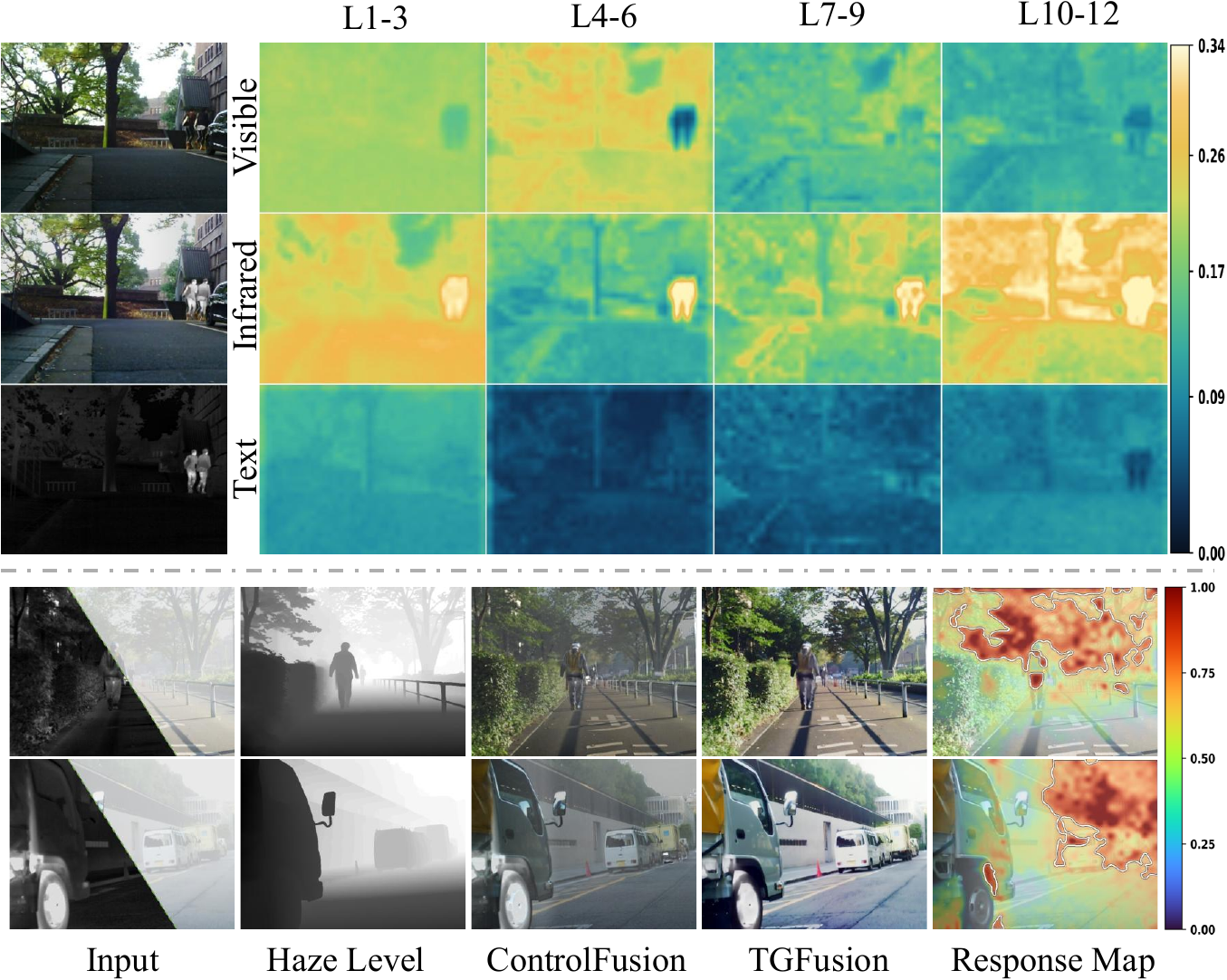}
    \caption{Visualization of semantic--visual interactions in TGFusion. Top: Stage-wise attention from the fusion stream to the VIS, IR, and Text streams. Bottom: Localization of the haze token and qualitative comparison with ControlFusion.}
    \label{fig:attention_response}
\end{figure}

\subsection{Fusion Comparison under Degradation Scenarios}
\label{sec:degraded_comparison} 

We further evaluate robustness under representative single-modality and compound degradations. As summarized in Table~\ref{tab:degradation}, TGFusion consistently ranks first or second, with obvious advantages in CLIP-IQA, MUSIQ, and TReS while maintaining competitive EN and SD. Compared with standard benchmarks, its gains are more pronounced under corrupted observations, indicating robustness to changes in modality reliability rather than to a specific type of degradation. Figure~\ref{fig:degradation} corroborates this trend. Despite degradation-specific pre-enhancement, competing methods still inherit restoration residuals. In particular, ControlFusion~\cite{tang2026controlfusion} retains veil-like haze and underexposure in the haze and low-light cases; under rain-haze and rain, visible rain streaks remain, and scene textures are degraded. TGFusion effectively suppresses these residual corruptions, including simultaneous low-light and stripe-noise interference, while recovering visible structures and preserving infrared-target saliency. The qualitative evidence agrees with the perceptual-metric gains, demonstrating more reliable aggregation of complementary information under complex degradations.
Please refer to Appendix~\ref{app:visual_ddl12} for more visual results.

\subsection{Analysis of Semantic--Visual Interaction}
\label{sec:semantic_visual}

We visualize the semantic--visual interactions within TGFusion. The upper panel of Figure~\ref{fig:attention_response} presents stage-wise attention from the fusion stream to the VIS, IR, and Text streams. Early blocks show broad responses across the three conditions, supporting initial inter-modal interaction. As features evolve, VIS attention increasingly traces scene structures, while IR attention concentrates on thermally salient targets. In later blocks, the Text stream also develops content-aware spatial responses. This broad-to-selective evolution suggests that joint interaction aligns text guidance with the emerging fusion objective, allowing text to support local refinement rather than remain a fixed global prior.

The lower panel of Figure~\ref{fig:attention_response} shows the response of the individual \emph{haze} token. Its high-response regions generally follow the spatial distribution of haze shown in the corresponding haze-level maps, particularly in heavily degraded areas. Compared with ControlFusion, which retains noticeable haze residuals in these regions, TGFusion more effectively suppresses haze in the corresponding high-response areas while preserving scene structures and infrared target saliency. The consistency among the haze distribution, token response, and fusion outcome indicates that the text stream can localize corrupted content and provide region-sensitive guidance for degradation suppression and reliable information aggregation across modalities.


\subsection{Semantic Segmentation Evaluation}
\label{sec:segmentation}

We further evaluate downstream semantic utility on MFNet~\citep{ha2017mfnet} by separately training SegNeXt~\citep{guo2022segnext} on infrared, visible, and each method's fused images under the same protocol. As reported in Table~\ref{tab:semantic-segmentation}, TGFusion achieves the highest mIoU and mAcc, reaching 71.83\% and 81.47\%, respectively. Qualitative comparisons in Appendix~\ref{app:segmentation} further show that effective information restoration and integration produce predictions more consistent with the ground truth, notably suppressing the false responses observed with competing methods. Detailed per-class IoU and accuracy results are also provided in the appendix~\ref{app:segmentation}.

\subsection{Ablation Study}
\label{sec:ablation}

\begin{figure}[!t]
    \centering
    \includegraphics[width=\linewidth]{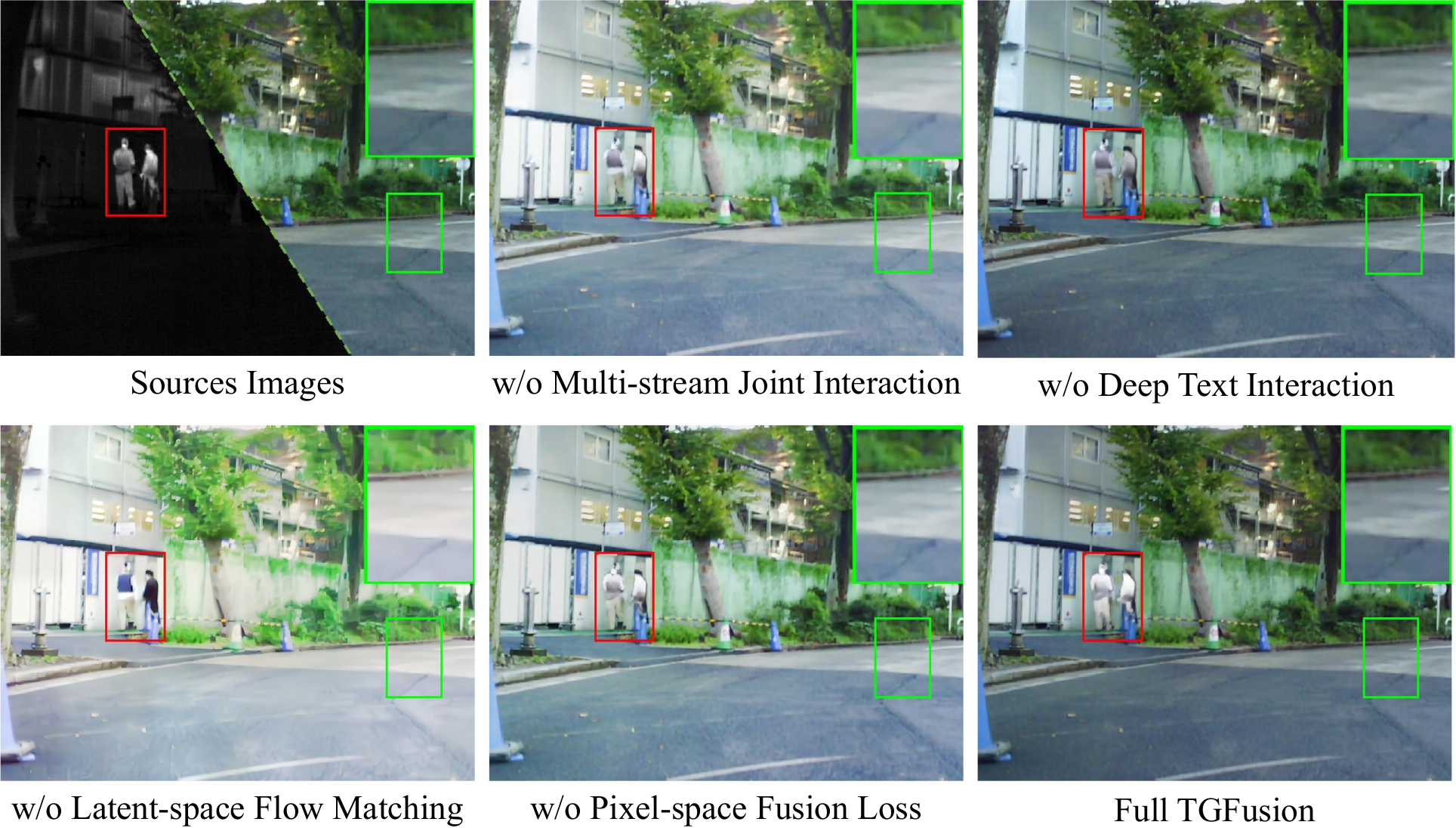}
    \caption{Qualitative comparison of network-component ablations on the standard MSRS dataset.}
    \label{fig:network_ablation_visual}
\end{figure}

\begin{table}[!t]
    \centering
    \renewcommand{\arraystretch}{1.0}
    {
    \small
    \setlength{\tabcolsep}{3pt}
    \begin{tabular}{@{}cccccc@{}}
        \toprule
        \textbf{Config.} & \textbf{EN} & \textbf{AG} &
        \textbf{CLIP-IQA} & \textbf{TReS} & \textbf{NIQE} \\
        \midrule
        \Rmnum{1} & 7.39 & 3.84 & 0.21 & 46.22 & 4.46 \\
        \Rmnum{2} & 7.15 & 3.71 & 0.17 & 36.73 & 5.01 \\
        \Rmnum{3} & 7.18 & 3.83 & \Second{0.25} & \Second{64.41} & 5.15 \\
        \Rmnum{4} & \Second{7.42} & \Best{4.37} & 0.20 & 46.52 & \Second{4.27} \\
        \Best{\Rmnum{5}} & \Best{7.49} & \Second{4.16} &
        \Best{0.28} & \Best{65.70} & \Best{3.67} \\
        \bottomrule
    \end{tabular}
    }
    \caption{Quantitative network-component ablation results on the standard MSRS dataset.}
    \label{tab:ablation}
\end{table}

As shown in Figure~\ref{fig:network_ablation_visual} and Table~\ref{tab:ablation}, replacing joint attention with independent stream-wise self-attention (\Rmnum{1}) reduces the relative saliency of infrared targets against bright background structures, with consistent degradation across all metrics. Removing the independent text stream (\Rmnum{2}) further weakens the emphasis on local thermal targets and produces the lowest EN, AG, CLIP-IQA, and TReS, demonstrating that global textual modulation alone cannot provide sufficient semantic--visual interaction. Performing flow matching directly in pixel space (\Rmnum{3}) causes severe brightness amplification, washed-out structures, and color artifacts. Although it achieves the second-best CLIP-IQA and TReS, its lower EN and AG together with the worst NIQE indicate inferior information preservation and visual naturalness. Removing the pixel-space fusion loss (\Rmnum{4}) introduces brightness drift and over-accentuates local structures, such as pedestrian contours and road cracks. Its highest AG therefore reflects stronger local gradients rather than balanced perceptual quality, as evidenced by the substantially lower CLIP-IQA and TReS. In contrast, the full model (\Rmnum{5}) preserves natural colors and visible structures while maintaining clear thermal-target saliency, achieving the best EN, CLIP-IQA, TReS, and NIQE. These results validate the complementary contributions of joint multi-stream interaction, the independent text stream, latent-space flow matching, and pixel-space fusion supervision.

\section{Conclusion}

This work demonstrates that textual priors can move beyond static global conditioning to actively support degradation-robust IVIF. By coupling progressive semantic--visual interaction with latent flow generation, TGFusion mitigates diverse degradations while retaining fine structural details and salient thermal targets. Results on standard benchmarks, single and compound degradations, and semantic segmentation confirm its robust fusion quality and downstream utility. These findings suggest a promising direction for reliable multimodal fusion under adverse imaging conditions.


\bibliography{main}

\clearpage
\appendix

\input{secs/AAAI2027_Supplementary}

\end{document}

%% file: secs/AAAI2027_Supplementary.tex
\begin{figure*}[!t]
    \centering
    \includegraphics[width=0.95\textwidth]{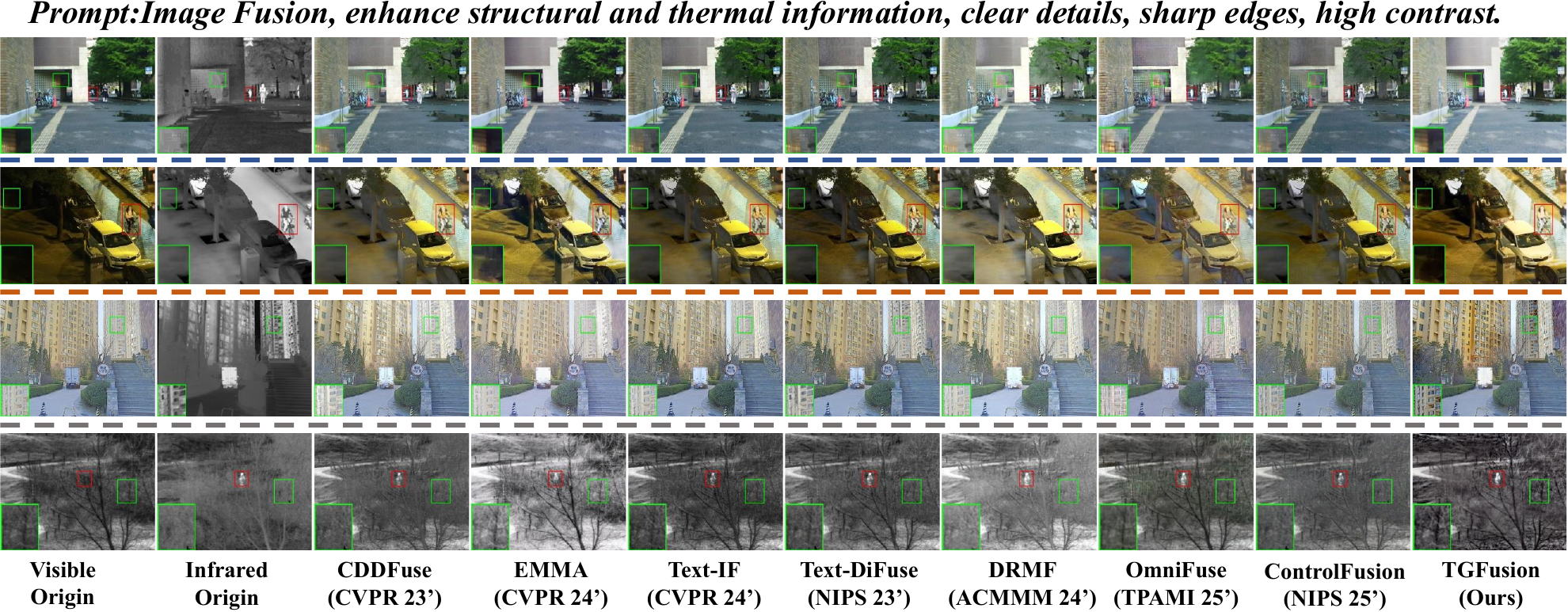}
    \caption{Qualitative comparisons on public benchmarks.}
    \label{fig:benchmark_qualitative}
\end{figure*}

\section*{Supplementary Contents}

This supplementary material provides further support for the methodology and experimental findings presented in the main paper. Specifically, it includes:

\begin{itemize}
    \item Detailed training and inference settings.
    \item Quantitative results on TNO.
    \item Qualitative comparisons on public benchmarks.
    \item Extended qualitative comparisons across diverse degradation scenarios.
    \item Detailed semantic segmentation results on MFNet, including qualitative comparisons and per-class IoU and accuracy evaluation results.
\end{itemize}

\section{Implementation Details}
\label{app:training_details}

TGFusion is trained on DDL-12~\citep{tang2026controlfusion} for 120 epochs using bf16 precision. Paired infrared and visible images undergo synchronized random horizontal and vertical flipping, followed by aligned random cropping to $256\times256$. The trainable parameters are optimized using AdamW with $\beta_1=0.9$, $\beta_2=0.999$, and a weight decay of $0.01$. The effective batch size is 32 across two GPUs. The learning rate is linearly warmed up to $1\times10^{-4}$ during the first 5 epochs and then cosine-decayed to $1\times10^{-6}$. Training is implemented in PyTorch 2.4.1 with CUDA 12.1 and conducted on two A100-SXM4-40GB GPUs for approximately three days.

During inference, the fused latent representation is generated using a 25-step Euler ODE solver with a classifier-free guidance scale of $3.0$.


\section{Quantitative Comparison on TNO}
\label{app:quantitative_tno}

Table~\ref{tab:tno} reports the complete quantitative results on TNO. TGFusion achieves the best EN, AG, and CLIP-IQA scores and ranks second in both TReS and NIQE, although it does not attain a leading CC score. This overall performance trend is consistent with the findings on the public benchmarks reported in the main paper, demonstrating TGFusion's competitive image fusion capability.


\section{Visual Comparison on Public Benchmarks}
\label{app:visual_public}

Figure~\ref{fig:benchmark_qualitative} presents qualitative comparisons on four public benchmarks. On MSRS, TGFusion preserves the salient thermal response of the pedestrian while maintaining illumination transitions and structural details in the surrounding dark regions. On LLVIP, EMMA exhibits regular grid artifacts despite its higher CC, whereas OmniFuse introduces conspicuous blockwise color distortions. TGFusion avoids these artifacts while retaining clear pedestrian, vehicle, and road structures. On M3FD, it alleviates the veil-like, low-contrast appearance observed in competing results and preserves both the salient vehicle response and clearer window and balcony structures. On TNO, TGFusion maintains a prominent small thermal target together with better-defined branches and ground textures. These comparisons further show that stronger source correlation can accompany the retention of degradations or artifacts. Overall, TGFusion achieves a more favorable balance among informative source-content preservation, reliable cue selection, and perceptual quality.


\section{Extended Qualitative Comparisons across Diverse Degradation Scenarios}
\label{app:visual_ddl12}

\begin{table}[!t]
    \centering
    \small
    \setlength{\tabcolsep}{2.5pt}
    \renewcommand{\arraystretch}{1.05}
    \begin{tabular}{@{}lcccccc@{}}
        \toprule
        \textbf{Method}
        & \textbf{EN}
        & \textbf{AG}
        & \textbf{CC}
        & \textbf{CLIP-IQA}
        & \textbf{TReS}
        & \textbf{NIQE} \\
        \midrule
        \textbf{CDDFuse}
        & 7.120 & 4.896 & \Best{0.513}
        & 0.262 & 27.718 & 4.346 \\

        \textbf{EMMA}
        & 7.158 & 4.760 & \Second{0.493}
        & \Second{0.287} & 28.026 & 5.206 \\

        \textbf{Text-IF}
        & 7.159 & \Second{4.917} & 0.484
        & 0.237 & 31.370 & \Best{3.923} \\

        \textbf{Text-DiFuse}
        & 7.145 & 4.115 & 0.472
        & 0.247 & 36.851 & 4.481 \\

        \textbf{DRMF}
        & \Second{7.195} & 4.462 & 0.384
        & 0.242 & 31.826 & 4.886 \\

        \textbf{OmniFuse}
        & 7.043 & 3.564 & 0.468
        & 0.221 & 36.401 & 6.189 \\

        \textbf{ControlFusion}
        & 6.977 & 4.479 & 0.488
        & 0.274 & \Best{53.162} & 4.524 \\

        \textbf{TGFusion}
        & \Best{7.315} & \Best{5.798} & 0.405
        & \Best{0.297} & \Second{36.967} & \Second{4.102} \\
        \bottomrule
    \end{tabular}
    \caption{Quantitative comparison on the TNO dataset.}
    \label{tab:tno}
\end{table}

\begin{figure*}[!t]
    \centering
    \includegraphics[width=0.95\textwidth]{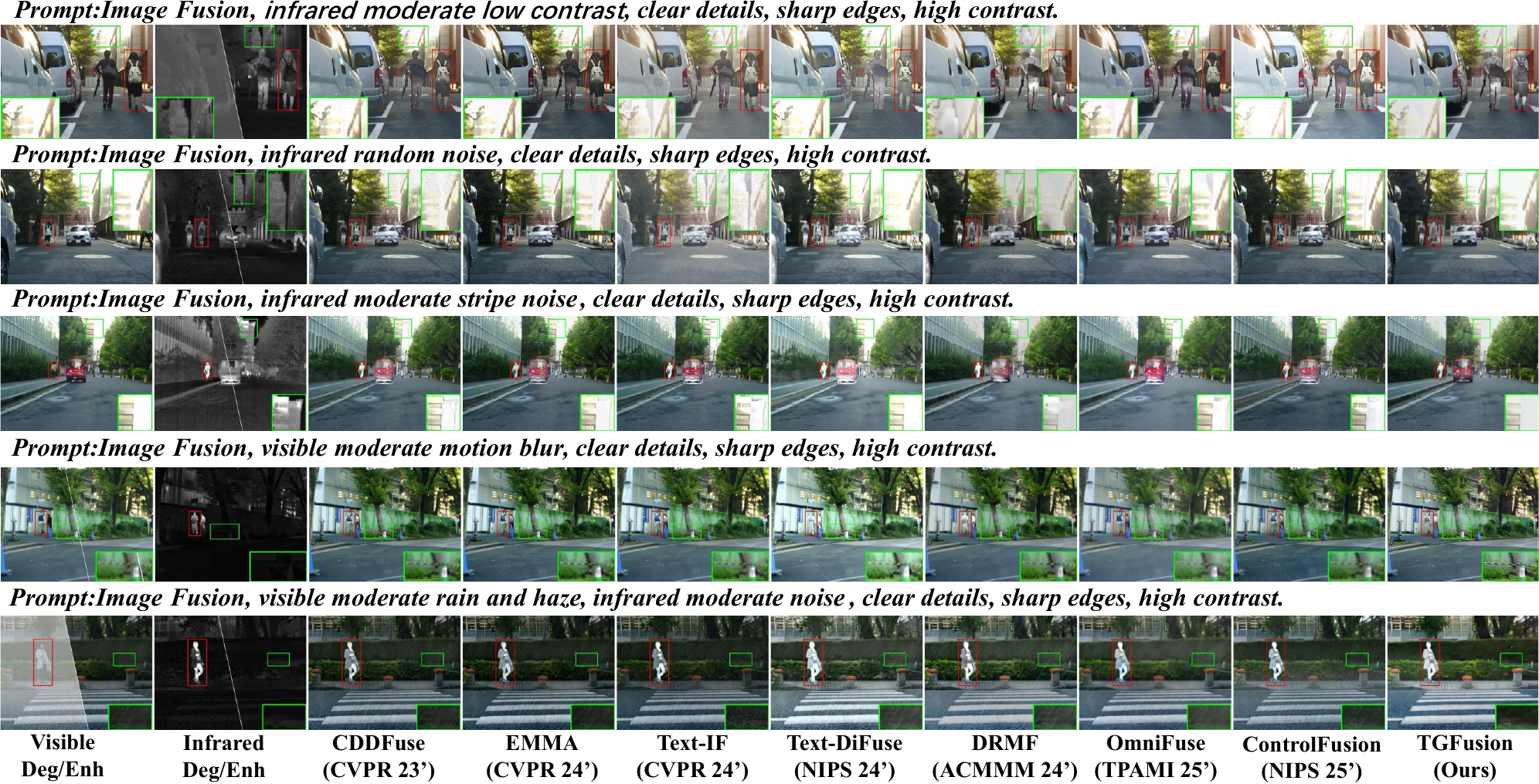}
    \caption{Additional qualitative comparisons under representative degradation scenarios from DDL-12~\citep{tang2026controlfusion}.}
    \label{fig:ddl12_more_qualitative}
\end{figure*}

\begin{figure*}[!t]
    \centering
    \includegraphics[width=0.80\textwidth]{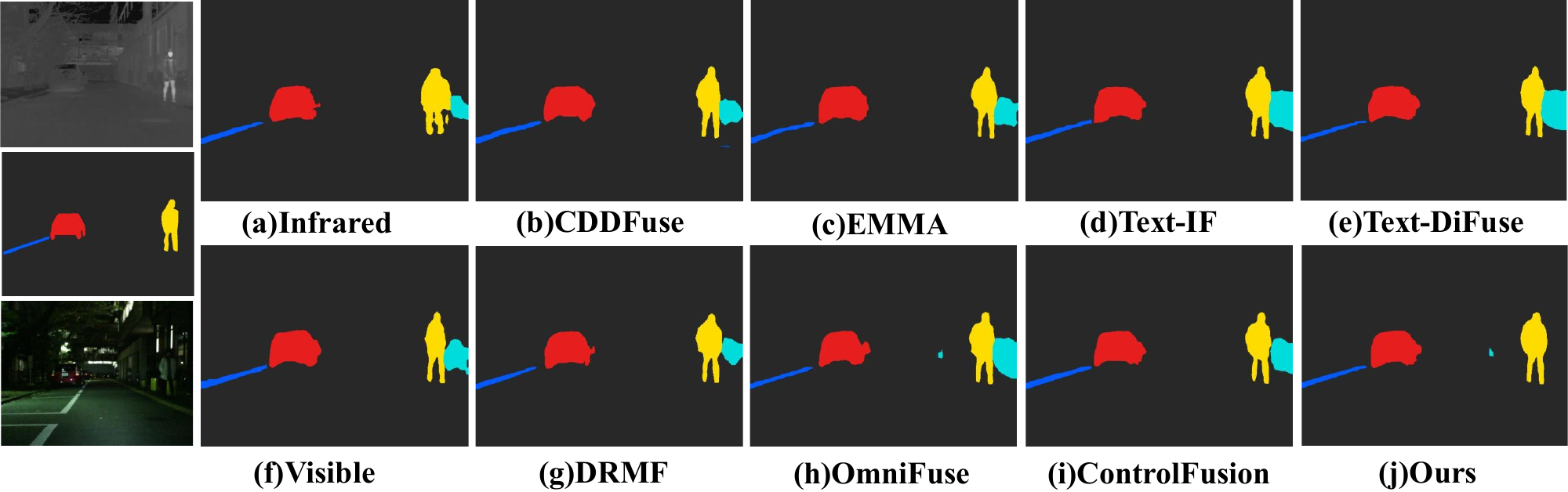}
    \caption{Qualitative semantic-segmentation comparison on MFNet~\citep{ha2017mfnet}.}
    \label{fig:semantic_segmentation}
\end{figure*}

Figure~\ref{fig:ddl12_more_qualitative} provides additional qualitative comparisons across representative modality-specific and cross-modal compound degradations. Competing methods often retain corrupted source cues, resulting in residual noise, rain streaks, brightness distortions, or excessive smoothing that weakens thermal saliency and structural details. In contrast, TGFusion more effectively limits degradation propagation while preserving salient targets and recovering scene content.

These results are consistent with the proposed design: structured degradation priors progressively interact with the visual streams through joint attention, facilitating the selection of reliable modality cues, while latent-space flow matching supports high-quality structure and texture recovery. Overall, TGFusion achieves a favorable balance among degradation mitigation, thermal-target preservation, and detail reconstruction across diverse and complex degradation scenarios.

\begin{table*}[!t]
    \centering
    \renewcommand{\arraystretch}{1.05}
    \setlength{\tabcolsep}{2.2pt}
    \small

    \begin{tabular}{@{}lcccccccccc@{}}
        \toprule
        \textbf{Class}
        & \textbf{Infrared}
        & \textbf{Visible}
        & \textbf{CDDFuse}
        & \textbf{EMMA}
        & \textbf{Text-IF}
        & \textbf{Text-DiFuse}
        & \textbf{DRMF}
        & \textbf{OmniFuse}
        & \textbf{ControlFusion}
        & \textbf{TGFusion} \\
        \midrule

        \multicolumn{11}{@{}l}{%
            \textit{Intersection over Union (IoU, \%)}
        } \\
        car
        & 87.60 & 89.56 & 89.41 & 89.16 & \Best{90.16}
        & 89.96 & 89.56 & 89.86 & \Second{89.99} & 89.94 \\

        person
        & 70.54 & 64.00 & 72.79 & 73.57 & \Best{75.11}
        & 74.37 & 73.34 & 72.21 & \Second{74.53} & 72.56 \\

        bike
        & 66.05 & 69.04 & 69.33 & 70.12 & 70.24
        & 70.24 & 70.18 & 68.38 & \Best{71.57} & \Second{70.58} \\

        curve
        & 52.41 & 56.50 & 58.91 & \Best{59.98} & 59.34
        & 59.56 & 58.08 & 55.78 & \Second{59.61} & 58.54 \\

        car stop
        & 66.87 & 73.74 & 73.09 & 73.56 & \Second{74.22}
        & 73.52 & 71.55 & 69.53 & 72.85 & \Best{76.95} \\

        guardrail
        & 56.64 & \Best{74.47} & 71.48 & 66.61 & 67.82
        & 70.74 & 67.97 & 64.43 & 65.70 & \Second{72.99} \\

        color cone
        & 60.02 & 66.63 & 63.21 & 64.73 & 66.52
        & \Best{69.12} & 61.62 & 66.34 & \Second{68.58} & 67.77 \\

        bump
        & 64.14 & 65.90 & \Best{68.60} & 67.32 & 62.33
        & 66.45 & \Second{68.04} & 66.42 & 66.00 & 65.28 \\

        \textbf{mIoU}$\uparrow$
        & 65.53 & 69.98 & 70.85 & 70.63 & 70.72
        & \Second{71.75} & 70.04 & 69.12 & 71.10 & \Best{71.83} \\

        \midrule

        \multicolumn{11}{@{}l}{%
            \textit{Per-class Accuracy (\%)}
        } \\
        car
        & 92.29 & 93.83 & 92.99 & 92.92 & \Best{94.35}
        & 94.09 & 93.98 & 93.69 & 93.85 & \Second{94.32} \\

        person
        & 81.61 & 74.00 & 83.48 & 83.62 & \Best{86.45}
        & 84.67 & 83.06 & 82.33 & \Second{85.28} & 82.93 \\

        bike
        & 76.63 & 77.64 & 77.81 & 78.37 & 78.22
        & \Second{79.49} & 77.93 & 76.24 & \Best{80.43} & 78.55 \\

        curve
        & 64.61 & 71.35 & 71.47 & \Best{75.20} & \Second{75.12}
        & 74.25 & 72.01 & 67.84 & 74.37 & 73.07 \\

        car stop
        & 77.42 & 81.50 & \Second{81.57} & 79.79 & 80.35
        & 80.75 & 77.99 & 75.87 & 80.78 & \Best{85.16} \\

        guardrail
        & 61.75 & \Best{84.58} & 77.07 & 70.05 & 74.27
        & 78.53 & \Second{79.37} & 75.05 & 79.23 & 79.24 \\

        color cone
        & 70.12 & 76.66 & 70.03 & 73.75 & \Best{80.77}
        & 77.67 & 73.21 & 76.21 & 77.78 & \Second{79.31} \\

        bump
        & 76.66 & \Best{82.82} & 78.97 & 76.85 & 72.50
        & 77.24 & \Second{80.97} & 80.65 & 76.89 & 79.15 \\

        \textbf{mAcc}$\uparrow$
        & 75.14 & 80.30 & 79.17 & 78.82 & 80.25
        & 80.84 & 79.82 & 78.49 & \Second{81.08} & \Best{81.47} \\

        \bottomrule
    \end{tabular}

    \caption{Per-class IoU and accuracy comparison for semantic segmentation on MFNet.}
    \label{tab:supp-semantic-segmentation}
    \label{tab:supp-semantic-segmentation-iou}
    \label{tab:supp-semantic-segmentation-acc}
\end{table*}

\section{Detailed Semantic Segmentation Results}
\label{app:segmentation}

Although the best per-class results are distributed among different methods, TGFusion achieves the highest mIoU and mAcc of 71.83 and 81.47, respectively. Its simultaneous advantages in region overlap and category-wise accuracy demonstrate balanced semantic utility rather than gains confined to isolated categories. As shown in Figure~\ref{fig:semantic_segmentation}, most competing methods produce a conspicuous bike response adjacent to the pedestrian despite its absence from the ground truth. TGFusion substantially suppresses this false response while preserving the principal car, person, and road-structure regions. Together, the quantitative and qualitative results indicate that TGFusion retains discriminative semantic information while reducing misleading cues, thereby supporting more reliable downstream segmentation.